# Sea ice detection using concurrent multispectral and synthetic aperture radar imagery.

Martin S J Rogers[a*], Maria Fox[a], Andrew Fleming[a], Louisa van Zeeland[b], Jeremy Wilkinson[a], and J. Scott Hosking[a,b]

[a]British Antarctic Survey, High Cross, Madingley Rd, Cambridge CB3 0ET

[b]The Alan Turing Institute, British Library, 96 Euston Rd., London NW1 2DB

Corresponding author*: marrog@bas.ac.uk

Author email addresses: marfox@bas.ac.uk, ahf@bas.ac.uk, lvanzeeland@turing.ac.uk, jpw28@bas.ac.uk, jask@bas.ac.uk

**Abstract**

Synthetic Aperture Radar (SAR) imagery is the primary data type used for sea ice mapping due to its spatiotemporal coverage and the ability to detect sea ice independent of cloud and lighting conditions. Automatic sea ice detection using SAR imagery remains problematic due to the presence of ambiguous signal and noise within the image. Conversely, ice and water are easily distinguishable using multispectral imagery (MSI), but in the polar regions the ocean's surface is often occluded by cloud or the sun may not appear above the horizon for many months. To address some of these limitations, this paper proposes a new tool trained using concurrent multispectral Visible and SAR imagery for sea Ice Detection (ViSual_IceD). ViSual_IceD is a convolution neural network (CNN) that builds on the classic U-Net architecture by containing two parallel encoder stages, enabling the fusion and concatenation of MSI and SAR imagery containing different spatial resolutions. The performance of ViSual_IceD is compared with U-Net models trained using concatenated MSI and SAR imagery as well as models trained exclusively on MSI or SAR imagery. ViSual_IceD outperforms the other networks, with a F1 score 1.30% points higher than the next best network, and results indicate that ViSual_IceD is selective in the image type it uses during image segmentation. Outputs from ViSual_IceD are compared to sea ice concentration products derived from the AMSR2 Passive Microwave (PMW) sensor. Results highlight how ViSual_IceD is a useful tool to use in conjunction with PMW data, particularly in coastal regions. As the spatial-temporal coverage of MSI and SAR imagery continues to increase, ViSual_IceD provides a new opportunity for robust, accurate sea ice coverage detection in polar regions.

# 1	Introduction

Sea ice affects oceanic-atmospheric fluxes of heat and gases (Haid and Timmermann, 2013), influences local climate conditions through changes to surface albedo (Riihela et al. 2021), provides habitat and breeding grounds for wildlife (Steiner et al. 2021), and potentially safeguards the fracture and calving of ice shelves by buttressing ice shelves and dissipating ocean wave energy (Christie et al. 2022). The rapid and accurate detection of sea ice is important for operational monitoring and for analysing large numbers of images to determine seasonal to decadal scale changes in polar ice in response to changing climatic conditions (Parkinson, 2019).

Remote sensing image analysis is well-suited to sea ice mapping due to the large spatial extent and dynamic nature of sea ice (Zakhvatkina et al. 2019). Sea ice coverage, investigated within this study, is defined as the total area of sea ice on the ocean's surface and differs from sea ice extent which is the area of ice with a sea ice concentration (SIC) greater than 15%. Sea ice coverage has historically been manually digitised from remote sensing imagery to produce sea-ice charts (National Snow and Ice Data Centre, 2023; Norwegian Meteorological Institute, 2023). The time-consuming, subjective nature of this exercise has motivated the development of automated tools for this task, particularly for shipping where accurate, low latency ice charts are necessary for safe navigation (Smith et al. 2022).

Three publicly available satellite image datasets have primarily been used for sea ice detection: (i) passive microwave (PMW) data (ii) multispectral imagery (MSI) and (iii) Synthetic Aperture Radar (SAR). PMW data products such as the Bremen SIC products derived from the Advanced Microwave Scanning Radiometer 2 (AMSR2) sensor and the Ocean and Sea Ice Satellite Application Facility (OSI SAF) SIC products derived from the Special Sensor Microwave Imager/Sounder (SSMIS) sensors provide unparalleled, multi-decadal, daily coverage of both polar regions (Spreen et al. 2008; Lavergne et al. 2019); however, its coarse spatial resolution precludes the precise identification of the ice-water interface. MSI platforms including the Moderate Resolution Imaging Spectroradiometer (MODIS) provide daily, $10^2$ m resolution, pan-Antarctic imagery of sea ice coverage (Roger et al. 2015). The spectral properties of ice and water are generally very distinct, but deleterious cloud cover and lighting conditions often preclude the analysis of MSI for sea ice detection (Lee et al. 2020). Radar imagery from the Sentinel-1 missions, referred to from here as SAR, captures information on sea ice coverage independent of cloud and lightning conditions, and is freely available with an approximate pixel spacing of 40 m in modes usually used for sea ice charting. However, this SAR imagery has a less frequent and lower spatial coverage and phenomena such as ocean waves and water melt on the ice surface can generate ambiguous textures in SAR imagery, making ice-water classification problematic (Stokholm et al. 2022). The benefits and limitations of MSI and SAR imagery justifies the development of tools to detect sea ice via multiple concurrent images, and to investigate the ability of these tools to utilise the benefits and overcome the limitations of MSI and SAR images respectively.

This paper investigates the viability of training a machine learning tool, namely a convolutional neural network (CNN), on concurrent MODIS MSI and Sentinel-1 SAR imagery to detect Antarctic sea ice coverage. The performance of two separate networks that combine the MSI and SAR imagery is explored:

(i) a traditional U-Net model trained using resized and fused MSI and SAR imagery (referred to from here as FuseNet); and

(ii) a model with novel CNN architecture that contains two parallel encoder phases (ViSual_IceD).

The relative performance of single and multiple encoder networks for image segmentation tasks has been compared in other domains including land cover classification (Marmanis et al. 2018) and object detection in natural red-green-blue imagery (Hazirbas et al. 2016), but the same analysis has not previously been performed for sea ice detection. The performance of these two multi-image networks is compared against two further U-Net models, the first trained exclusively on MSI scenes (MSI network) and the other trained exclusively using SAR imagery (SAR network). The relative importance of the MSI and SAR imagery for sea ice detection is investigated using the explainable AI technique, permute and predict. The outputs from the best performing network are compared to a time series of SIC values derived from the AMSR2 sensor in the South Bellingshausen Sea.

All studies addressing the same issue have developed and applied tools exclusively to Arctic imagery. Compared with the Arctic, Antarctic sea ice backscatter properties can differ because it is generally younger, thinner, contains fewer melt ponds, has a greater number of icebergs, has less ridging and is located at lower latitudes (Zakhvatkina et al. 2019). It is also more prevalent in the Antarctic for snow fall to depress the sea ice, causing it to flood and refreeze forming 'snow ice'. Detailed understanding of Antarctic sea ice conditions is necessary because its extent has not exhibited the same linear reduction over the past 40 years as in the Arctic (Parkinson, 2019). These points highlight the necessity of developing separate networks with appropriate training datasets to detect sea ice coverage at high resolution in the Antarctic and Arctic.

## 2 Machine learning applications to sea ice detection

Numerous machine learning tools have been developed to detect sea ice coverage or classify sea ice type, with most previous research using object-based classifiers such as CNN (Yu et al. 2023). A CNN is a form of machine learning that stacks successive layers of convolution and pooling to identify scale-invariant features within the original input image. Internal weights within the CNN architecture connect the input image to successive features maps computed by the convolutional kernels, and activation functions are applied to the weights to enable the model to learn non-linear functions. During the training stage, epochs of feed-forward passes and back propagation successively update the internal weights contained within the model, see Gu et al. (2018) for greater detail on CNN operations and recent advances. Numerous CNN architectures exist, with the U-Net architecture most widely used for sea ice detection in remote sensing imagery (Ronneberger et al. 2015).

The U-Net architecture is an end-to-end CNN where an output layer is produced at the same resolution and extent as the original input image (Ronneberger et al. 2015). The encoder stage contains the alternate convolution and pooling layers to extract features from the image. The decoder consists of layers of deconvolution or up-sampling to reconstruct the segmented output layer using the image features. The outputs from the encoder stage are passed via skip connections to the decoder stage to aid image segmentation. Recent advancements in the application of U-Net models to sea ice include increasing the receptive field or the size of the neighbourhood of pixels the CNN considers when classifying each output pixel (Stokholm et al. 2022); employing an ensemble of U-Net models (Wang and Li 2021) and altering the batch size, the number of images the model is simultaneously trained on (Boulze et al. 2020). Despite these advances, most papers highlight the potential of fusing multiple data sources prior to model training to improve network performance.

Models trained using multiple data sources commonly utilise SAR data due to the rich textual information the imagery provides (Han et al. 2021). Malmgren-Hansen et al. (2021) trained a U-Net model using SAR imagery as input and subsequently injected SIC data derived from the AMSR2 sensor into a deeper layer of the model, reducing noise in segmented outputs. Gelis et al. (2021) concatenated SAR imagery with reanalysis data of wind speed to address the change in backscatter properties of water driven by wind strength and angle. The inclusion of this wind layer was found to have negligible benefit on the performance of the CNN but was retained in case it provided more information to the CNN on the radiometric ambiguities between sea ice and open water. PMW and reanalysis data have a coarse spatial resolution, restricting the ability to resolve low-level features such as edges with high precision. This highlights the need to investigate the fusion of other data sources containing similar spatial resolutions.

Models using concurrent MSI and SAR imagery remain uncommon, despite the finer spatial resolution and pixel spacing of both data sources. Konig et al. (2021) combined these two image types to classify sea ice type and thickness. Initially, the authors calculated ratios of reflectance values from different channels in MSI scenes and a Support Vector Machines classification algorithm was independently applied to the corresponding SAR imagery. The outputs of the two classification tasks were combined via multiplication and principal component analysis to generate a final ice classification map. Han et al. (2021) extracted features individually from hyperspectral MSI and SAR imagery using two separate deep learning models before fusing spectral and spatial joint features to classify SIC in Hudson Bay, Canada. Both projects highlight the potential of training a CNN using concurrent MSI and SAR imagery, but only applied their trained network to an individual test site and used a cloud mask prior to model training, meaning sea ice information is not provided by the networks in locations where cloud occludes the MSI scene. Therefore, further investigation is required to train a CNN using concurrent

MSI and SAR imagery where a cloud mask is not applied and to determine if the CNN can generalise to segment sea ice coverage in a range of new locations.

# 3   Methods
## 3.1   Image data sources and preprocessing

Copernicus Sentinel-1 A and B satellites capture C-Band (5.5 cm wavelength) Synthetic Aperture Radar (SAR) imagery (European Space Agency, 2023). This study acquired single-polarised, HH, SAR images in the Extra Wide (EW) swath mode with an approximate swath width of 400 km and an approximate pixel spacing of 40 m. Single polarisation images were exclusively used due to the failure of the Sentinel 1B satellite and subsequent reduction in dual polarisation imagery availability in the Antarctic region. Imagery was collated from across West Antarctica with the majority acquired in the Lazarev Sea, Weddell Sea, Bellingshausen Sea, Amundsen Sea, Ross Sea, and Antarctic Peninsula (Figure 1). Due to lighting conditions, all imagery was captured in the austral summer between October - March inclusive from 2016 - 2022.

Sentinel-1 SAR imagery was sourced and preprocessed using an automated programmatic method developed using the SentiNel Application Program (SNAP) tool (Zuhlke et al. 2015). To address ambiguities in SAR imagery deriving from the elevated backscatter values of water in areas of low (steep) incidence angle, this study compared the performance of ViSual_IceD when the model was trained using three different SAR processing methods: (i) orbit files were applied to all images prior to thermal noise removal, sigma0 calibration and terrain correction, (ii) after applying the steps in (i), radiometric terrain flattening was applied, which accounts for changes in incidence angle caused by topography, and (iii) after applying the steps in (i), a novel method was used which normalises the backscatter values by incidence angle (Evans et al. 2023). Appendix 2 summarises the findings of this comparative investigation, most notably that the addition of the steps outlined in (ii) and (iii) reduced the number of false positives (water misclassified as ice) in areas of low (steep) incidence angle, but the image-wide performance of ViSual_IceD reduced. This is attributed to a reduction in the contrast of backscatter values for ice and water in other parts of the image when applying the additional processing steps listed in (ii) and (iii). Subsequently, the models trained within this project utilised SAR imagery processed using the procedures outlined in (i).

NASA's Moderate Resolution Imaging Spectroradiometer (MODIS) satellite provides MSI scenes at 250 m spatial resolution. This study used a combination of visible and short-wave infrared (SWIR) MODIS-Terra Corrected Reflectance imagery wavebands, namely band 3 (459 – 479 nm), band 6 (1628 - 1652nm) and band 7 (2105–2155 nm) (MODIS Science Team 2017), where the initial values within each band range [0, 255]. This combination was chosen because it is well-suited to distinguishing between ice, cloud and water. Ice strongly reflects visible radiation but absorbs most SWIR, whereas water and cloud absorb and reflect all three bands respectively (Roger et al. 2015). Each band used had an initial resolution of 500 m but was sharpened to 250 m using the corresponding panchromatic band prior to extraction from the MODIS repository.

A publicly available interface was developed to automatically extract the corresponding MSI scene for each SAR image from NASA's Global Imagery Browse Services (GIBS) repository (Rogers, 2022). To collate spatially and temporally concurrent MSI and SAR imagery, the acquisition time and bounding coordinates of each SAR image were initially collected. MSI and SAR image pairs acquired > 8 hours apart from each other were immediately discarded. Three factors were considered when deciding to retain the remaining MSI-SAR image pairs: (i) the time difference between the acquisition of the two images, (ii) the location of image capture and (iii) the identification of major discrepancies between the location of the major ice-water boundary in the image pairs, determined via manual inspection during the generation of refernce images (Section 3.2). Ice drift models were referred to when identifying the maximum allowable time lag between MSI and SAR image pairs (Kimura 2004; Farooq et al. 2020). For locations with the greatest rates of ice drift, such as the North Weddell Sea, images pairs were only retained when the MSI and SAR scenes were acquired within 1-2 hours of each other, whereas lag times between image acquisition was up to 8 hours in locations where sea ice drift velocities are lower, including the Weddell Sea and Western Peninsula.

Manual inspection of all remaining image pairs was subsequently conducted in GIS software, QGIS, to determine whether any major discrepancies, > 720 m or 3 CNN output image pixels, in the position of the major ice-water interface between the two images remained. A minimum zoom level of 1:250000 was used, enabling a detailed comparison of the location of the ice-water edge in the MSI and SAR image pairs.

### 3.2   Reference sea ice classification layer generation

To train a supervised CNN, corresponding reference sea ice classification layers of ice (1) and water (0) were generated. Figure 2 outlines the steps used to generate the reference sea ice classification for the ViSual_IceD, FuseNet and MSI-only networks. A threshold was first applied to the visible band of the MSI scene to produce an initial binary layer of ice (including icebergs) and water (Figure 2 (c)). After manual comparison of the outputs derived from applying different threshold values to the visible band, a threshold value of 70 was selected to ensure that the majority of mixed ice-water pixels were classified as ice, reducing underrepresentation of sea ice extent in the marginal ice zone. To correct for cloud erroneously labelled as ice, an additional threshold was applied to the SWIR band to distinguish ice from cloud (Figure 2 (d)). A low threshold value of 20 was applied to the SWIR band to ensure all cloud was identified in the image. The corresponding SAR image was used to manually digitise all areas of open water and major in-ice water features, such as large leads and polynyas, under cloud using QGIS software (Figure 2 (e)). The manually digitised polygons were applied as a mask to the initial binary layer to produce the final reference sea ice classification (Figure 2 (f)). Note, for the SAR-only network, the reference imagery was exclusively generated by manually digitising the entire SAR scene in QGIS. The corresponding MSI scene was referred to during digitisation where ambiguous signals or low incidence angles made it difficult to decipher the location of the major ice-water boundary.

MSI and SAR images and their corresponding reference sea ice classification were combined to generate two separate sets of image pairs: SAR-reference and MSI-reference, and one set of image triplets: SAR-MSI-reference. The SAR-reference dataset was produced at 80 m resolution, whereas the MSI-reference and SAR-MSI-reference datasets were generated at 240 m resolution. See Figure 3 for a summary of the processing workflow employed in this study.

### 3.3  Data augmentation and paired image creation

Data augmentation techniques were employed to increase the size of the image training dataset. All MSI and SAR images were resized to 240 m and 80 m resolution respectively and normalised to [-1, 1] to increase the speed of CNN training. Resized MSI and SAR images were subset into patches with width and heights of 240 × 240 and 720 × 720 pixels respectively. These dimensions ensured that MSI and SAR patches had the same footprint. To further ensure spatial overlap between the MSI and SAR patches, every patch pair was subset using coordinates rather than image row and column values. The corresponding reference sea ice classification was also subset into patches with the same extent. Any patches containing NaN values at the boundary of the original SAR images were discarded. All patches were flipped along the vertical axis and rotated by 90, 180 and 270 degrees. This resulted in a total of > 80,000 SAR-MSI-reference patch triplets.

Fewer than 20% of patches contained both ice and water, meaning greater than 80% of the patches from the corresponding reference sea ice classification were labelled as all water (0) or all ice (1). To reduce the likelihood of the CNN segmenting every future unseen image to be exclusively ice or exclusively water, CNN performance was compared when trained on datasets containing different proportions of patches containing ice-water boundaries. Accordingly, three image datasets were produced: (i) one containing all patches generated (all images), (ii) one containing only the patches with an ice-water boundary (edge images) and (iii) one containing all the edge patches, and the same number of patches containing all water and all ice pixels (equal images). Concretely, equal images contained three times the number of patches as edge images.

### 3.4  Convolutional neural network architecture design

Four separate CNN were trained to detect sea ice coverage within this study (Figure 3). Three separate CNNs with U-Net architectures were developed and trained using (i) SAR imagery (SAR network) (ii)

MSI scenes (MSI network) or (iii) fused MSI and SAR imagery (FuseNet). To produce the FuseNet training dataset, the SAR image was resized to the same, coarser resolution as the MSI scene and concatenated with two of the three bands in the MSI scene (visible and SWIR). The final CNN developed in this study contained a novel architecture, ViSual_IceD, that contains two parallel encoder phases for the two image types (Figure 4). Three CNNs, ViSual_IceD, FuseNet and the MSI network were trained to generate outputs at 240 m resolution, whereas the SAR network produced outputs at 80 m resolution.

Numerous modifications were made to the original U-Net architecture: (i) A batch normalisation layer was added between each set of convolutional layers to reduce CNN training time; (ii) an additional convolutional layer with stride = 2 was used instead of the 2 × 2 max pooling layer to allow the CNN to learn its own pooling function; and (iii) deconvolution layers were replaced by up-sampling layers within the decoder stage to reduce the presence of 'checkerboard' features in the CNN outputs (Odena et al. 2016). The above-mentioned model improvements were also employed in the ViSual_IceD network.

To account for the differences in the spatial resolution between the two image types in the ViSual_IceD architecture, the SAR encoder stage contains an additional set of convolutional and downsizing layers than the MSI encoder stage (Figure 4). The SAR encoder stage contains a 3 × 3 max pooling layer after the first set of convolutional layers to create a feature layer with height and width of 240 pixels. This ensures all MSI and SAR feature layers in subsequent layers of the model have the same height and width, allowing these feature layers to be concatenated and provided via skip connections to the decoder stage (Figure 4). The decoder stage correspondingly contains four sets of up-scaling and convolutional layers, resulting in an output image being produced with the same dimensions and spatial resolution as the input SAR imagery.

## 3.5 Convolutional neural network training

For every CNN, the augmented datasets described in Section 3.3. were split into 80% training and 20% validation data. The Rectified Linear Unit (ReLU) activation function was used in each model to learn non-linear relations and the cross-entropy loss function was employed to update internal architecture weights.

Hyperparameters such as learning rate, batch size and input dataset have a large influence on CNN performance, but individual testing of all combinations of hyperparameter values can be time consuming. To overcome this issue, the WandB package was employed to run many CNN training iterations using different hyperparameter values. WandB uses a Bayesian optimisation algorithm to efficiently identify combinations of hyperparameter values that reduce validation loss values (Snoek et al. 2012). Table 1 outlines the hyperparameters and the range of values tested within this study as well as the best hyperparameter values identified for each CNN. Memory capacity prevented the use of batch sizes greater than 32. For the transfer learning hyperparameter, the network performance was tested using the pretrained ImageNet weights (Stanford Vision Lab, 2016). A class balance constant of 0.1 and 10 encouraged the detection of water and ice respectively. To prevent overfitting, the maximum number of epochs during model training was set at 100 epochs and an early stopping criterion was employed to stop training if loss did not improve over five epochs.

All CNN training and hyperparameter tuning was implemented in Python 3.8 using the Keras library with Tensorflow backend. Training was run on a Nvidia Quadro P4000 graphical processing unit (GPU) and the maximum run time for training any individual CNN was 8.5 hrs.

## 3.6 Convolutional neural network testing

To validate the performance of the four CNN tested within this investigation, four pixel-based evaluation metrics were used: binary accuracy (Equation 1), user accuracy or precision (Equation 2), producer accuracy or recall (Equation 3) and F-Score (Equation 4). Each network was applied to 40 previously unseen West Antarctic test images. The same preprocessing steps outlined in Section 3.1. were applied to each test image.

$$B_A = \frac{T_P + T_N}{T_P + T_N + F_P + F_N} \qquad (1)$$

$$U_A = \frac{T_P}{T_P + F_P} \qquad (2)$$

$$P_A = \frac{T_P}{T_P + F_N} \qquad (3)$$

$$F_1 = \frac{U_A + P_A}{U_A \times P_A} \qquad (4)$$

where $B_A$, $U_A$, $P_A$ and $F_1$ are the binary accuracy, user accuracy, producer accuracy and F-Score values, respectively. $T_P$ = true positive and $T_N$ = true negative, each corresponding to correctly classified pixels and $F_P$ = false positives and $F_N$ = false negative, each corresponding to incorrectly classified pixels. A lower $P_A$ corresponds to a greater number of pixels misclassified as ice and a lower $U_A$ is caused by more pixels being incorrectly classified as water. The metrics were applied to the entire test set, and then the performance of ViSual_IceD was compared between three separate regions of Antarctica: (i) Amundsen and Bellingshausen Sea, (ii) Antarctic Peninsula and West Weddell Sea, and (iii) East Weddell Sea and Indian Ocean. To assess the impact of low incidence angle in the SAR image on network performance, the evaluation metrics were also exclusively applied to pixels in the ViSual_IceD outputs corresponding to areas of low incidence angle in the SAR image. Low incidence angle is defined here as $< 25^0$ which corresponds to the maximum incidence angle found in the first swath of each of the SAR test images.

All output images from the four networks contained values [0,1], corresponding to the confidence of the network that a particular pixel should be classified as ice. To apply the validation metrics, a threshold was first applied to all output imagery to generate a binary ice (1) and water (0) mask. Sensitivity analysis of the threshold value used was conducted by applying every threshold value between 0.1 and 0.9 with 0.1 intervals.

When applying the trained networks to images larger than the 720 x 720 pixel patch size, the original test images were split into patches of this size, with an overlap of 120 pixels between each adjacent patch. The trained network was then applied to each patch and the patches were subsequently merged to reconstruct an output image with the same dimensions as the original image. To mitigate edge effects when applying the trained tool, each patch was cropped by 60 pixels at each boundary, also meaning there was no overlap with adjacent prediction patches.

The non-parametric McNemar test was applied to determine whether there was a statistically significant difference between the locations where ViSual_IceD and the other three networks incorrectly classified the test images (Equation 5). For each test image, the outputs from ViSual_IceD and the other networks were compared to the reference image to generate binary True/False maps, corresponding to locations where the networks correctly or incorrectly classified that pixel.

$$M = \frac{(T_V - F_V)^2}{(T_V + F_V)} \tag{5}$$

where $T_v$ corresponds to locations where the ViSual_IceD classification was correct and the other network was incorrect, and $F_v$ corresponds to locations where the ViSual_IceD classification was incorrect and the classification from the other network was correct. A higher value corresponds to larger disagreements in the locations where the networks were incorrect. The null hypothesis states that there is no difference in the locations where the two networks incorrectly classify the image. The null hypothesis was rejected when a p-value < 0.05 was produced, meaning there is a statistically significant difference in the locations where ViSual_IceD and the other networks incorrectly classified the image.

### 3.7  Permute and predict.

The permute and predict method was employed to ascertain information on the relative importance of the MSI and SAR image for the multi-image networks. The purpose of permute and predict is not to generate viable network outputs, but to improve understanding of how the trained network derives its ice-water charts. Initially, the pre-trained multi-image network was applied to a concurrent MSI-SAR image pair. A different SAR image, acquired from a different date and location, was then paired with the original MSI scene. A binary ice-water mask was then generated when the multi-image network was applied to the original MSI - permuted SAR pair (permuted output).

To calculate the impact of permuting the SAR image, the permuted output was subtracted from the original output using pixel-wise calculations. A value close to 0 (1) corresponds to small (large) changes in the original and permuted outputs respectively. To determine whether cloud cover affected the relative importance of the two image types, the above-mentioned pixel-wise calculations were compared in locations where cloud was and was not present in the MSI scene. Cloud locations were identified by applying a threshold to the SWIR band to generate a cloud mask after permuted image segmentation. The magnitude of change between the original and permuted outputs was subsequently compared for pixels in and out of cloud cover. This process was carried out on 10 different original MSI-SAR pairs using 10 permuted images each time, meaning this permute and predict experiment was conducted a total of 100 times.

### 3.8 Comparing CNN outputs to passive microwave data

The best performing network, ViSual_IceD, was subsequently applied to a time series of images of the Carroll Inlet, South Bellingshausen Sea, to detect the reduction in sea ice during the austral summer 2018 - 2019 (see inset Figure 1). This site was selected because it is of navigational importance as the coastal topography of Stange Sound allows offload of fuel and science equipment by ship for onward transport into the Antarctic continent. Sea ice conditions are highly variable and accurate sea ice information is an important part of planning and ship operations.

ViSual_IceD outputs were compared to concurrent sea ice concentration (SIC) maps previously derived from applying the ARTIST Sea Ice (ASI) algorithm to microwave radiometer data captured by the Advanced Microwave Scanning Radiometer 2 sensor (AMSR2 product) (Spreen et al. 2008). The AMSR2 product provides daily, pan-Antarctic SIC layers at a 6.25 km spatial resolution. To allow direct comparison between the two data products, ViSual_IceD outputs were downscaled to 6.25 km resolution by convolving a mean kernel filter over the initial ViSual_IceD binary ice-water output. This kernel had a spatial footprint and stride of 6.25 km that enabled the network to calculate the proportion of pixels classified as ice by ViSual_IceD in each AMSR2 product grid square, equating to SIC. A land mask was applied to the ViSual_IceD data, setting all land pixels to Not a Number (NaN), enabling the mean filter kernel to only consider ocean pixels when calculating the proportion of pixels classified as ice.

# 4  Results
## 4.1  Relative network performance

Figures 5, 6 and 7 provide a visual comparison of outputs produced by the four networks tested within this project that were trained using the hyperparameters defined in Table 1. Figures 5 and 6 provide examples of the original outputs from the networks scaled [0,1] and Figure 7 shows network outputs after a threshold has been applied to generate a binary ice-water mask. The outputs from the networks show the confidence of the pixel corresponding to ice. A value close to 1 (0) corresponds to the network classifying the pixel with a high confidence as ice (water) respectively. Pixel values close to 0.5 correspond to locations where the network has a low confidence that the pixel is either ice or water.

Binary accuracy, $U_A$, $P_A$ and F1 scores produced by the four networks on the 40 test images are provided in Table 2. ViSual_IceD had the highest mean binary accuracy (0.942), $U_A$ (0.983) and F1 score (0.972) of the four networks whilst FuseNet had a higher mean $P_A$ (0.961). The higher $U_A$ and F1 score achieved by ViSual_IceD shows that the network produces a low number of false positive results where water is misclassified as ice. The higher false positive rate of FuseNet meant the network did not detect many in-ice water features like ViSual_IceD. The lower $P_A$ score of ViSual_IceD corresponds to a higher number of false negatives results, where the network misclassifies ice as water. ViSual_IceD performance was consistent between the West Weddell and Amundsen and Bellingshausen Sea test areas where F1 scores of 0.972 and 0.979 respectively were recorded, but the network produced inferior $P_A$ and F1 values in the East Weddell Sea and Antarctic Peninsula test area (F1 = 0.931). Further, ViSual_IceD $U_A$ was lower in areas corresponding to low incidence angle in the SAR image ($\mu$ = 0.901) compared with results for entire images ($\mu$ = 0.983).

The multi-image networks outperformed the networks trained exclusively on MSI or SAR imagery for all four validation metrics. The MSI network achieved $U_A$ scores similar to ViSual_IceD (0.955 and 0.983 respectively) but had an inferior $P_A$ (0.913 compared to 0.951). The SAR network achieved the lowest validation metric scores of the four networks (F1 = 0.927). The SAR network also achieved the lowest $U_A$ (0.898), commonly generating many false positive results as can be seen at the bottom of Figure 7 (f). The McNemar test revealed a statistically significant difference between the locations where ViSual_IceD and the other models incorrectly classified the test image (p-value range 0.001-0.04).

An important finding was that the networks trained on concurrent MSI and SAR imagery produced outputs that classify pixels as ice or water with greater confidence. Across all test images, the proportion of output pixels with a value 0.05 < $y$ < 0.95 was 8% for ViSual_IceD, 12% for FuseNet, 21% for the MSI network, and 29% for the SAR network. It is desirable to have a low proportion of pixels valued 0.05 < y < 0.95, as this corresponds to a high proportion of pixels being confidently classified as either ice or water by the network. Concordantly, the performance metrics of ViSual_IceD was the least affected by the choice of threshold value applied to generate a binary ice-water mask. The binary accuracy changed by < 4% when using a threshold of 0.1 verses 0.9, but interestingly the changes were greater for the other networks (see Appendix 1 for full analysis of the impact of threshold value selection on network performance). The best threshold value to use for ViSual_IceD and FuseNet was 0.9, whereas it was 0.7 for the MSI network and 0.6 for the SAR network.

An example of low confidence in the classification made by the MSI network is shown in the blue box in Figure 5 (e), whereas ViSual_IceD can resolve sea ice coverage with high network confidence in the same location (Figure 5 (c)). The SAR and MSI networks both have low confidence when classifying the boundary between the iceberg and open water in the green square in the bottom left of the image (Figure 5 (e) & (f)). In comparison, the ViSual_IceD network discriminates between the location of the iceberg and open water with high confidence (Figure 5 (c)).

## 4.2  Relative importance of different data sources

To test the relative importance of the MSI scene verses SAR image, the SAR image was permuted and the magnitude of difference in ViSual_IceD output using the original SAR image (original output) and permuted SAR image (permuted output) calculated (Figure 8). Positive (negative) values correspond to the original output having a higher (lower) confidence than the permuted output of the pixel being sea ice. The mean change in pixel value across all permuted image outputs was $\mu = 0.17$ with a standard deviation, $\sigma = 0.28$. In both cloudy and clear conditions, the difference between original and permuted output tended to be positive, meaning that permuted outputs values were lower, and permuting the SAR image increased the number of pixels classified as water (Figure 8). The lower proportion of change values between $-0.75 < y < -0.25$ and $0.25 < y < 0.75$ is the result of a high proportion of ViSual_IceD output pixels being close to 0 or 1, meaning change values would be closer to -1, 0 or 1.

In the ViSual_IceD network, cloud cover affected the magnitude of change in the pixel value when permuting the SAR image. Greater differences in permuted and original ViSual_IceD output values were recorded under cloudy than clear conditions (Figure 8). Mean pixel value differences between permuted and original ice marks were $\mu = 0.32$ and $\mu = 0.07$ under cloudy and clear conditions respectively. The standard deviation of differences in pixel values was also greater in cloudy conditions, $\sigma = 0.42$, than in clear conditions, $\sigma = 0.15$. The greater use of the SAR scene during image segmentation when the MSI scene is occluded by cloud is also demonstrated visually in Figure 9, where sea ice conditions are completely occluded by cloud in the MSI scene, and the sea ice coverage classified by ViSual_IceD closely resembles that contained within the SAR image. In comparison, cloud cover had less of an impact on the magnitude of change in pixel value when permuting the SAR image in the FuseNet network. For this network, mean pixel value differences between permuted and original ice marks were $\mu = 0.11$ and $\mu = 0.09$ under cloudy and clear conditions respectively.

## 4.3 Sea ice variability in Carroll Inlet

Figure 10 provides a comparison of a time series of sea ice conditions in the Carroll Inlet depicted by the AMSR2 product at 6.25 km, binary ice-water masks produced by ViSual_IceD, and SIC values generated by downscaling ViSual_IceD outputs to 6.25 km resolution. The AMSR2 product and ViSual_IceD datasets both show similar global patterns in sea ice condition over this period, including a consistent reduction in sea ice coverage between each time stamp and the Stange Sound coastline first becoming ice free in the same location in late January 2019 (Figure 10 (d), (i) and (n)).

ViSual_IceD provides more detailed information on the ice-water boundary and identifies the location of individual floes that have detached from the main region of sea ice adjacent to Stange Sound. The finer resolution of the ViSual_IceD outputs enables the network to detect in-ice water features adjacent to Stange Sound that are not detectable within the AMSR2 product. Downscaled ViSual_IceD consistently show sea ice extent and concentration to be lower in the Carroll Inlet than in the concurrent AMSR2 product.

## 5 Discussion

### 5.1 Relative network performance

ViSual_IceD is the first example of a CNN trained to detect sea ice coverage in concurrent MODIS multispectral imagery (MSI) and Sentinel 1 synthetic aperture radar (SAR) imagery. A comparison of evaluation metrics show that networks trained on multiple image sources can outperform analogous networks trained on individual image types (Table 2). This study investigated different methods via which a CNN could be trained on concurrent MSI and SAR imagery:

(i)  concatenate or fuse both image types prior to input into the CNN with single encoder (FuseNet); and

(ii)  create a new CNN architecture, ViSual_IceD, with dual encoder (Figure 4).

ViSual_IceD was shown to outperform FuseNet, with higher binary accuracy, $U_A$ and F1 scores (Table 2), and the results from the McNemar test showed a statistically significant difference in the location of incorrectly classified pixels between the ViSual_IceD and FuseNet networks (Equation 5). The superior performance of ViSual_IceD is primarily attributed to the network's ability to utilise the most informative image type in different circumstances. In cloud free conditions, ViSual_IceD almost exclusively uses the MSI scene to detect sea ice conditions. This is shown by the small differences between original and permuted outputs classification values indicating that the image is classified independent of ice conditions depicted within the SAR image ($\mu = 0.07$, Figure 8). Larger differences between the original and permuted output values under cloudy conditions indicates that changes in sea ice conditions in the SAR image influences network segmentation ($\mu = 0.32$). In comparison, FuseNet permuted and original output values were more similar, indicating that the network primarily classifies using the MSI scene in all situations ($\mu = 0.11$). This inability for FuseNet to select the most informative image type during segmentation is disadvantageous and provides reasoning for its inferior performance compared to ViSual_IceD. It is noted that permuted output values could be caused by the generation of unrealistic combinations of pixel values that the network has not had exposure to during training (Fisher et al. 2019). Further, it was not possible to derive meaningful results from permuting the MSI scene and retaining the same SAR image, as permuting the MSI scene affected both the cloud and sea ice conditions present, overriding any observable effects on ViSual_IceD outputs caused by the presence of noise and ambiguous signal in the SAR image. These considerations mean the permute and predict methods indicate, but do not prove, this selective use of imagery by ViSual_IceD.

Visual inspection of outputs from ViSual_IceD provide further evidence of the selective use of SAR imagery by ViSual_IceD in cloudy conditions (e.g. Figure 9). ViSual_IceD outputs closely resemble the sea ice coverage in SAR images when cloud completely occludes the ocean surface in the corresponding MSI scene. Further, in the blue box in Figure 5 (c), ViSual_IceD provides a high confidence output that closely resembles the sea ice conditions contained within the SAR. In comparison, the MSI network classifies this part of the image with low confidence because the network has no information pertaining to the ocean surface (blue box Figure 5 (e)). The visual comparison of network outputs further support assertions deduced from the permute and predict outputs that the ViSual_IceD network can selectively choose to use the MSI or SAR scene dependent upon cloud conditions.

The detection of sea ice in locations occluded by haze and cloud by ViSual_IceD is a key advancement on previous work. Previous studies have used concurrent optical and radar satellite imagery to detect sea ice extent or concentration but have applied a cloud mask prior to network training and application (Han et al. 2021; Konig et al. 2021). The extent of cloud masks is dependent upon the algorithm used and difficulties persist in robustly identifying some cloud types such as thin cirrus clouds or haze from land signatures (Foga et al. 2017; Lee et al. 2020). As such, partial cloud occlusion remains a limitation even when cloud masks are applied. Further, a cloud mask can occlude a substantial proportion of the ocean's surface meaning information of sea ice coverage in many areas remain unobtainable. This highlights the potential advantages of training ViSual_IceD to have exposure to many cloud conditions to make it transferable to many new test circumstances.

Where the SAR image contained ambiguous signal, noise, or higher backscatter due to lower incidence angle, ViSual_IceD used the MSI scene during image segmentation [Figure 5 (c) and (f) and Figure 7 (c), (e) and (f)]. The SAR-only network incorrectly classified water as ice at the bottom of Figure 7 (f), where wind-roughened water and a low incidence angle were present, whereas ViSual_IceD and FuseNet networks correctly identified these areas as ocean. The mean $U_A$ of ViSual_IceD was lower at areas of low incidence angle ($U_A = 0.901$), compared to the entire test image set ($U_A = 0.983$). This is attributed to both cloud and low incidence angle in some MSI-SAR image pairs; however, visual inspection of the outputs demonstrate an accurate segmentation of ice from water in most situations, despite the lower $U_A$ (Figure 5 (c) and 6 (c)). Whilst the combination of MSI and SAR imagery shows promise in overcoming unavoidable noise and ambiguous signal, efforts to make the network more robust to all circumstances remains necessary.

Correct image preprocessing methods are key to overcome noise and ambiguous signals in SAR imagery. These impacts have previously been mitigated by increasing the receptive field of the network, altering the SAR image preprocessing chain (Stokholm et al. 2022), running ensemble models (Wang and Li, 2021), or modifying the CNN architecture (Gelis et al. 2021). Noise and ambiguous signals are discussed as persistent limitations in SAR-only networks. This paper investigates the impact of incidence angle normalisation (Evans et al. 2023) and radiometric terrain flattening to mitigate the effects of these features. Both methods reduced the overall performance of ViSual_IceD, attributed to a reduction in contrast in backscatter values for water and ice (Appendix 2). Future research is required to identify the best SAR image preprocessing techniques to handle the effects of low incidence angle in the input SAR image (Mäkynen and Karvonen, 2017; Mahmud et al. 2018).

This study produced a comprehensive training set consisting of 80,000 image patches containing a diverse range of sea ice, cloud, and wind conditions across the West Antarctic. Alongside increasing the size of the reference dataset via manual digitisation, future work should investigate the use of semi-supervised methods and bootstrapping methods that have shown great promise in detecting features in SAR images such as sea ice and glacial crevasses (Surawy-Stepney et al. 2023; Han et al. 2020) despite being trained on a relatively small training dataset. Both methods train deep learning tools using both labelled and unlabelled data, leveraging the vast amount of information contained within unlabelled imagery to improve the performance of the networks (Han et al. 2020). Future research should investigate the potential of using these methods for detecting sea ice extent when using multiple concurrent image sources.

Despite the benefits that using concurrent imagery provides, some limitations in the ViSual_IceD network persist, for example the lower $P_A$ value, corresponding to ice misclassified as water. This is attributed to the presence of complex cloud boundaries and cloud shadows that reduce surface reflectance making sea ice more spectrally similar to water (Figure 6 (c); Lee et al. 2020). This finding is consistent with previous CNN applications that have incorrectly classified ice as water at cloud boundaries (Hoffman et al. 2021). Lower $P_A$ values are also attributed to modifications in the backscatter properties of sea ice as temperatures near $0^O C$ as the snow layer on top of the ice become saturated, which can lead to the formation of brine-influenced snow-ice on the sea ice surface when new snow or rain is added (Casey et al. 2016; Paul et al. 2017). The lower $P_A$ values recorded in the East Weddell Sea and Antarctic Peninsula are attributed to these snow and ice properties that exist in warmer conditions that are more prevalent in these regions of lower latitude. The challenges of segmenting an image containing heterogeneous ice conditions and complex cloud boundaries could be further addressed by increasing the size and diversity of the training dataset to reduce the risk of overfitting, and investigating the use of other image sources, including altimetry data or SAR imagery with HV polarisation, which are less susceptible to these limitations.

### 5.2  Limitations of using concurrent imagery

A key consideration when training a machine learning tool using multiple image types to detect sea ice coverage is the drift in sea ice position between the acquisition of the two images. Contemporary ice drift models have determined that mean austral summer ice drift velocity in most parts of the Weddell Sea, Western Antarctic Peninsula, and along many coastlines are 1- 6 cm/s or 36 - 216 m/hr (Kimura 2004; Farooq et al. 2020). Conversely, ice drift rates can exceed 15 cm/s or 540 m/hr in the Drake Passage, the North Weddell Sea and Ross Sea, particularly where sea ice concentrations are lower, and wind has a greater influence on ice drift velocity (Farooq et al. 2020). This study defined 480 m or two ViSual_IceD output pixels as an acceptable drift distance between image acquisition. Using this information, only images acquired within 1 hour of each other are suitable in the more dynamic regions, but lag times between image capture can exceed 2 - 8 hours in the Weddell Sea or Western Peninsula. This is analogous to the findings of Konig et al. (2021), who determined 3 - 5 hours to be acceptable time lag when using combined Sentinel-3 multispectral and Sentinel-1 SAR imagery to map sea ice concentration. Further work could investigate whether providing information on image time lag during model training provides benefit, although within-image heterogeneity in ice-drift velocities may limit the information the model gains from this.

The continued increase in the spatio-temporal coverage of MSI and SAR imagery in both polar regions will act to reduce the circumstances where MSI and SAR imagery with an acceptable time lag is not available. MODIS MSI scenes are captured daily, and the Visible Infrared Imaging Radiometer Suite (VIIRS) instrument observes the earth's surface twice per day, providing imagery that is radiometrically similar to MODIS (USGS, 2023). In favourable locations along the Antarctic Peninsula and Western Weddell Sea, SAR imagery is acquired more than weekly from the Sentinel-1 satellite platforms (Liang et al. 2022). Coverage in polar regions will increase with the impending launch of Sentinel 1C and the Sentinel-1 high level operation plan contains objectives to revisit all Antarctic sea ice every three days (Wilson et al. 2021). Radarsat satellites capture radar imagery at the same resolution, C-band frequency, and polarisation as Sentinel 1 (Morena et al. 2004), and further investigation could look at the performance of a tool trained using both image types. Other existing or planned missions, such as the SAR Observation and Communications Satellite (SAOCOM), and the NASA-ISRO Synthetic Aperture Radar (NISAR), provide radar imagery with different frequencies and polarisation. Future work could investigate the training of a CNN for sea ice detection using multi-frequency and polarimetric radar imagery as inputs. Future improvements in the coverage of both MSI and SAR imagery in polar regions will increase the likelihood of obtaining concurrent imagery acquired within an acceptable time lag. This will help to reduce the limitations and potentially support the benefits of training machine learning tools on multiple image types to detect sea ice conditions.

### 5.3 Future applications of ViSual_IceD

The automated nature of ViSual_IceD provides potential to conduct extensive comparisons of SIC values derived from this tool and those acquired from passive microwave (PMW) data. Despite the widescale use of PMW data to derive SIC values, limitations remain in the algorithms used to convert the PMW brightness measurements into SIC values (Andersen et al. 2006; Kern et al. 2019). This study compared a time series of SIC values derived from microwave radiometer data captured by the Advanced Microwave Scanning Radiometer 2 sensor (AMSR2) and downscaled ViSual_IceD outputs in the Carroll Inlet (Figure 10). Both timeseries show the same dominant pattern of persistent sea ice retreat, but in many locations, SIC values were higher in the AMSR2 product than the ViSual_IceD outputs. This is attributed to the footprint of the coarse pixels from the AMSR2 product spanning coastal water and adjacent snow-covered land that is characterised by larger PMW emissions (Maa and Kaleschke 2010). This land contamination artificially increases the SIC value derived by the PMW dataset, whereas more robust SIC values are derived from ViSual_IceD in coastal regions where a fine detailed land mask can be applied. There is potential to use ViSual_IceD or similar tools trained on finer resolution imagery at a larger scale to refine algorithms used to derive SIC values from PMW radiometer data in these coastal regions. More broadly, comparisons of PMW SIC values and those derived from finer resolution imagery have been conducted in the Arctic, identifying underestimations in SIC derived from PMW data, particularly in summer months (Heinrichs et al. 2006; Wang and Li 2021; Shi et al. 2021). This highlights the potential to conduct this type of research in the Antarctic. The average processing time required to generate a new ViSual_IceD output is 4.5 minutes, providing promise that this type of analysis could be conducted at large spatio-temporal scales, as well as making ViSual_IceD a viable tool for generating sea ice charts for operational purposes.

The availability of MSI imagery is dependent upon lighting conditions during the winter and polar nights. Image availability varies across the Southern Ocean, although lighting conditions have a lower effect on image availability in the Antarctic than the Arctic because sea ice exists at lower latitudes. Daily imagery of the entire Southern Ocean is captured via the MODIS sensor between September - April. At lower latitudes ($< 64^0$), primarily along the Antarctic Peninsula, Weddell Sea, and Ross Sea, MSI scenes are available every other day in May, early June, late July and August, with no imagery available between mid-June and mid-July. At higher latitudes, one image is captured every three days in May and August and no imagery is available in June or July. ViSual_IceD provides promise in being a useful tool to supplement pre-existing PMW-derived SIC images between the Austral Spring-Autumn, coinciding with periods when the greatest uncertainties in PMW-derived SIC values exist and virtually all Antarctic marine navigation takes place.

Alongside downscaling the outputs of ViSual_IceD to make direct comparisons with PMW data, there is also opportunity to produce finer spatial resolution SIC charts. Detailed SIC derived from applying deep learning tools to finer resolution imagery has been conducted in the Arctic (Wang and Li 2021), but little investigation into this in the Antarctic has been achieved. Finer resolution SIC charts would be particularly useful in coastal regions to inform coastal habitat mapping (Jenouvrier et al. 2006; Rode et al. 2015), to aid in-ice navigation when polar vessels need access to research stations and other land-based features (Smith et al. 2022), and for aiding the activities of local communities in the Arctic. SAR imagery does not have the same spatio-temporal coverage as PMW data, as such there is potential to use PMW data to understand sea ice dynamics in both the Arctic and Antarctic, and use ViSual_IceD outputs to provide more detailed information for specific areas of interest when data is available. Additionally, it is noted that ViSual_IceD aggregates the classification of sea ice and icebergs into one class. The outputs of this network could be combined with models that focus exclusively on distinguishing between sea ice and icebergs to provide a more comprehensive depiction of sea ice conditions (Evans et al. 2023; Koo et al. 2023).

# 6   Conclusion

This study trained and applied a convolutional neural network with novel dual-encoder architecture, ViSual_IceD, to detect sea ice coverage using concurrent multispectral and synthetic aperture radar (SAR) imagery. The use of concurrent imagery as model input was found to improve network performance compared with networks trained exclusively on one image type. The results of the explainable AI technique permute and predict also indicate that ViSual_IceD selects the most informative image type during image segmentation. Concretely, the network primarily utilises the multispectral scene, which is less susceptible to noise, but uses the corresponding SAR image when the multispectral image is occluded by cloud.

ViSual_IceD outputs from the Carroll Inlet, South Bellingshausen Sea, were converted to sea ice concentration (SIC) layers and compared to SIC values derived from the AMSR2 passive microwave sensor. ViSual_IceD was shown accurately derive SIC values in coastal regions, whereas the AMSR2 product overestimated SIC where the footprint of the pixel overlapped ocean and snow-covered land. This provides promise that this tool or similar could be applied at large scales in conjunction with PMW derived SIC charts for operational and scientific purposes.

# 7 Acknowledgements

This work was funded by the Natural Environment Research Council (NERC). MR, MF and AF were supported under the British Antarctic Survey (BAS) core Polar Science for Planet Earth science strategy, and LVZ, JW and JSH were funded by the NERC programme grant "Drivers and Effects of Fluctuations in sea Ice in the ANTarctic" (DEFIANT, NERC References: NE/W004747/1, NE/W004755/1). This work is also a contribution to the Artificial Intelligence for Earth Observation Accelerator Project (Reference: 4000131401/20/I-NS). This work was partially supported by EPSRC Grant EP/Y028880/1 and The Alan Turing Institute. Thanks to members of The Mapping and Geographic Information Centre (MAGIC) at the British Antarctic Survey for producing some of the manually referenced sea ice classifications for the SAR-only network.

**Figure captions**

Figure 1: Location of test images visualised in this paper. Inset shows Carroll Inlet test area discussed in Section 3.9.

Figure 2: Workflow used to generate reference sea ice classification, (a) original MSI scene, (b) original SAR image, (c) simple threshold method to distinguish water (blue) from ice and cloud (yellow) (d) MSI scene classified into ice (yellow), water (blue) and cloud (red) using an additional threshold method, (e) manual digitisation of areas of water under cloud (green), (f) output reference sea ice classification.

Figure 3: Summary of the processing workflow employed in this study to train four convolutional neural networks.

Figure 4: ViSual_IceD architecture.

Figure 5: Network outputs for the East Weddell Sea test area. (a) Original MSI scene, (b) Original SAR image, outputs from (c) ViSual_IceD, (d) FuseNet, (e) MSI network and (f) SAR network respectively. Colour ramp for panels (c) to (f) shows relative confidence of pixel corresponding to ice. Values close to 1 correspond to a high confidence of ice (yellow), values close to 0 correspond to a high confidence of water (black) and values close to 0.5 correspond to areas of low classification confidence (orange, pink and purple). Blue and green boxes denote areas where ViSual_IceD classifies the image with higher confidence than the MSI or SAR network.

Figure 6: Network outputs for Stancomb Wills test area. (a) Original MSI scene, (b) Original SAR image, outputs from (c) ViSual_IceD, (d) FuseNet, (e) MSI network and (f) SAR network respectively. Colour ramp for panels (c) to (f) described in Figure 5.

Figure 7: Binary ice-water masks derived from each network for the Antarctic Peninsula test area. (a) Original MSI scene, (b) Original SAR image, outputs from (c) ViSual_IceD, (d) FuseNet, (e) MSI network and (f) SAR network respectively.

Figure 8: Difference in output classification pixel values when ViSual_IceD used the original and permuted SAR image as input. Positive (negative) values correspond to the original classification having a higher (lower) confidence of being ice. Pixel values are split by those in cloudy (grey) and clear sky (blue) conditions.

Figure 9: Zoomed-in subset of (a) original MSI scene from the South Bellingshausen Sea test area where the ocean surface is not discernible due to cloud cover, (b) corresponding SAR image and (c) ViSual_IceD output.

Figure 10: Comparison of sea ice conditions derived from various sources in the Carroll Inlet between December 2018 and February 2019. (a) - (e) AMSR2 SIC product values, (f) - (j) ViSual_IceD binary ice-water classification, and (k) - (o) ViSual_IceD outputs downscaled to the same resolution as the AMSR2 product and converted to SIC values.

**Table captions**

Table 1: Hyperparameter tuning values.

Table 2: Comparison of network performance.